%% file: terse_paper.tex
\documentclass{article}
\usepackage{spconf,amsmath,graphicx}

\title{Personalized Speech recognition on mobile devices}

\name{Ian McGraw, Rohit Prabhavalkar, Raziel Alvarez, Montse Gonzalez Arenas, Kanishka Rao,}
\nameplus{ David Rybach, Ouais Alsharif, Ha\c{s}im Sak, Alexander Gruenstein, Fran\c{c}oise Beaufays, Carolina Parada}

\address{Google Inc. \\
{\scriptsize \tt \{imcgraw,prabhavalkar,raziel,montse,kanishkarao,rybach,oalsha,hasim,alexgru,fsb,carolinap\}@google.com}}

\begin{document}
\maketitle
\ninept
\input{abstract}
\input{intro}
\input{baseline}
\input{am}

\input{lm}
\input{decoder}

\input{context}
\input{footprint}

\input{conclusion}
\input{acknowledgements}
\bibliographystyle{IEEEbib}
\bibliography{refs}
\end{document}

%% file: abstract.tex
\begin{abstract}
We describe a large vocabulary speech recognition system that
is accurate, has low latency, and yet has a small enough memory
and computational footprint to run faster than real-time on a Nexus 5
Android smartphone.  We employ a quantized Long Short-Term Memory
(LSTM) acoustic model trained with connectionist temporal
classification (CTC) to directly predict phoneme targets, and further
reduce its memory footprint using an SVD-based compression scheme.
Additionally, we minimize our memory footprint by using a single language model
for both dictation and voice command domains, constructed using Bayesian
interpolation.
Finally, in order to properly handle device-specific information,
such as proper names and other context-dependent information, we
inject vocabulary items into the decoder graph and bias the
language model on-the-fly. Our system achieves 13.5\% word error rate on an
open-ended dictation task, running with a median speed that is seven times
faster than real-time.
\end{abstract}

%% file: intro.tex
\section{Introduction}
\label{sec:intro}
Speech recognition for dictation, search, and voice commands has become a
standard feature on smartphones and wearable devices.  The vast majority of the
literature devoted to improving accuracy for these tasks assumes that speech
recognition will be run in datacenters on powerful servers.  However, despite
increases in speed and the availability of mobile internet, speech recognition
requests frequently have high latency, or even completely fail, due to
unreliable or unavailable network connections.  An embedded speech recognition
system that runs locally on a mobile device is more reliable and can have lower
latency; however, it must be accurate and must not consume significant memory or
computational resources.

In this paper we extend previous work that used quantized deep neural
networks (DNNs) and on-the-fly language model rescoring to achieve
real-time performance on modern smartphones~\cite{lei2013}.  We
demonstrate that given similar size and computation constraints,
we achieve large improvements in word error rate (WER) performance
and latency by employing Long
Short-Term Memory (LSTM) recurrent neural networks (RNNs), trained with
connectionist temporal classification (CTC)~\cite{GravesFernandesGomezEtAl06}
and state-level minimum Bayes risk (sMBR)~\cite{Kingsbury:2009} techniques.
LSTMs are made small and fast enough for embedded speech recognition
by quantizing parameters to 8 bits, by using context
independent (CI) phone outputs instead of more numerous context
dependent (CD) phone outputs, and by using Singular Value Decomposition
(SVD) compression~\cite{XueLiGong13, prabhavalkar2016}.

SVD has elsewhere been shown to be effective for speech processing
tasks~\cite{XueLiGong13,chen2015,NakkiranAlvarezPrabhavalkarEtAl15} as have
structured transforms~\cite{sindhwani2015} and low-rank matrix
factorizations~\cite{SainathKingsburySindhwaniEtAl13}. Vector quantization has
also been shown to significantly reduce model size with only small accuracy
losses~\cite{wang2015}, however it is unclear whether this algorithm
can be implemented in a computationally efficient manner while minimizing
runtime memory footprint.  Such parameter reduction techniques
have generally been applied to DNNs and not RNNs.  For embedded speech
recognition, some authors have avoided RNNs citing increased computational
costs and instead evaluated methods for transferring knowledge from RNNs to
DNNs~\cite{chan2015}.

We present results in two very different domains: dictation and voice commands.
To keep the disk space requirements of the system small, we experiment with
language model interpolation techniques that enable us to effectively share a
single model across both domains.  In particular, we demonstrate how the
application of bayesian interpolation out-performs simple linear interpolation
for these tasks.

Finally, we explore using language model personalization techniques to improve
voice command and dictation accuracy.  Many voice commands can be completed and
executed on a device without a network connection, or can easily be queued up to
be completed over an unreliable or slow network connection later in the
background.  For example, a command such as ``Send an email message to Darnica
Cumberland: can we reschedule?'' can be transcribed by an embedded speech
recognition system and executed later without a perceptual difference to the
user.  Accurate transcription, however, requires integrating personal
information such as the contact name ``Darnica Cumberland'' into the language
model. We demonstrate that the vocabulary injection and on-the-fly language
model biasing techniques from~\cite{aleksicIcassp2015,hall2015} can
significantly improve accuracy without significant adverse computational
overhead.

The remainder of this paper is organized as follows. We summarize the baseline
sytem in Section~\ref{sec:baseline}. Section~\ref{sec:am} describes our
techniques to build a small but accurate acoustic model, Section~\ref{sec:lm}
describes our LM training procedure and the interpolation techniques used in our
system, Section~\ref{sec:decoder} describes the decoder.
Section~\ref{sec:context} describes how we handle context or device-specific
information, and finally Section~\ref{sec:footprint} summarizes the footprint
of our system. Conclusions are presented in Section~\ref{sec:conclusion}.

%% file: baseline.tex
\section{Baseline System}
\label{sec:baseline}

We model our baseline system after the embedded speech recognition system
presented in~\cite{lei2013}. Instead of using a standard feed-forward DNN,
however, we use deep LSTM models which have been shown to achieve
state-of-the-art results on large-scale speech recognition
tasks~\cite{SakSeniorBeaufays14, SakSeniorRaoEtAl15a, SakSeniorRaoEtAl15b}.
The LSTM architecture of our baseline consists of three hidden layers
with 850 LSTM cells in each.  We make use of a recurrent projection layer
as described in ~\cite{SakSeniorBeaufays14} of size 450 for each of hidden
layers. This LSTM is trained
to predict 2,000 CD states, analogous to the system described in
~\cite{lei2013}. This system is also trained to optimize the standard (CE)
criterion on the training set, with the output labels delayed by 5
frames~\cite{SakSeniorBeaufays14}.

The input features are 40-dimensional log mel-filterbank energies calculated on
a 25ms window every 10ms. Unlike in~\cite{lei2013}, where frames are stacked to
provide right and left context to the net, we rely on the LSTM's memory
capabilities and supply only one frame every 10ms as input. This
model was trained to optimize the standard cross-entropy (CE) criterion on the
training set described in Section~\ref{sec:mtr}, with frame-level labels
derived from a larger system.

The language model presented in this work also follows along the lines of
~\cite{lei2013}.  The vocabulary size is restricted to 64K so that an index
into the lexicon only requires 16-bits of storage.  The small decoder graph is
constructed from a tiny LM containing 70K n-grams (almost entirely of
unigrams).  During decoding the partial paths are rescored on-the-fly with a
large LM containing roughly 1.5M n-grams.  This rescoring LM is made extremely
compact using the LOUDS~\cite{sorensen2011} compression mechanism. More details
of the LM can be found in Section~\ref{sec:lm}.

%% file: am.tex
\section{On-device Acoustic Modeling}
\label{sec:am}

In this section we describe an LSTM configuration that can successfully be
deployed to a mobile device and contrast this with the baseline system described
in Section~\ref{sec:baseline}.

In particular, the LSTM architecture that we investigate is a \emph{CTC
model}~\cite{SakSeniorRaoEtAl15a, SakSeniorRaoEtAl15b}: the system consists of
five hidden layers with 500 LSTM cells in each, that predict 41 context
independent (CI) phoneme targets plus an additional ``blank'' target that can
be hypothesized if the system is unsure of the identity of the phoneme at the
current frame. The system is trained to optimize the connectionist temporal
classification (CTC) criterion~\cite{GravesFernandesGomezEtAl06} as described
in~\cite{SakSeniorRaoEtAl15a, SakSeniorRaoEtAl15b}.

Similar to the baseline, we use standard 40-dimensional log mel-filterbank
energies over the 8Khz range, computed every 10ms on 25ms of input speech. In
order to stabilize CTC training, our CTC models use the strategy proposed
in~\cite{SakSeniorRaoEtAl15b}: we stack together 8 consecutive frames (7 frames
of right context) and only present every third stacked frame as input to the
network. In addition to stabilizing CTC training, this has the additional
benefit of speeding up computation since the network is only evaluated every
30ms.

\subsection{AM Experiments}
\label{sec:mtr}
Our AMs are trained on ~3M hand-transcribed anonymized utterances extracted from
Google voice search traffic (approximately 2,000 hours). All models in our work
are trained using distributed asynchronous stochastic gradient descent
(ASGD)~\cite{DeanCorradoMongaEtAl12}.  In order to improve robustness to noise
and reverberation, we generate ``multi-style" training data by synthetically
distorting each training utterance using a room simulator with a virtual noise
source, to generate 20 distorted versions of each utterance.  Noise samples are
extracted from YouTube videos and environmental recordings of daily events.

Results in this section are reported on a set of 13.3K anonymized utterances in
the domain of open-ended dictation extracted from Google traffic.  The LM used
in these experiments was described in Section~\ref{sec:baseline} and detailed
further in Section~\ref{sec:lm}. We benchmark our systems to determine runtime
speed by decoding a subset of 100 utterances on a Nexus 5 smartphone which
contains a 2.26 GHz quad-core CPU and 2 GB of RAM. We report median real-time
factors (RT50) on our test set. Our results are presented in
Table~\ref{tbl:am-results}.

\begin{table}
\begin{center}
\begin{tabular}{|l|c|c|c|c|}
\hline
AM Setup & WER & Params & Size & RT50\\
\hline
\hline
LSTM 2,000 CD States & 23.4 & 9.9M & 39.4 MB & 2.94 \\
\hline
LSTM CTC CI Phones & 19.4 & 9.7M & 38.8 MB & 0.64 \\
+ sMBR & 15.1 & 9.7M & 38.8 MB & 0.65 \\
+ SVD Compression & 14.8 & 3M & 11.9 MB & 0.22 \\
+ adaptation & 12.9 & 3M & 11.9 MB  & 0.22 \\
+ quantization & 13.5 & 3M & 3 MB & 0.14 \\
\hline
LSTM CTC (Server-size) & 11.3 & 20.1M & 80.4 MB & - \\
\hline
\end{tabular}
\end{center}
\vspace{-5pt}
\caption{Word Error Rates (\%) on an open-ended dictation task, evaluating
various acoustic models, using the same language model described in
Section~\ref{sec:lm}, along with median RT factor.}
\label{tbl:am-results}
\end{table}

As can be seen in Table~\ref{tbl:am-results}, and consistent with previous
work~\cite{SakSeniorRaoEtAl15a}, the CTC-trained LSTM model that predicts CI
phones outperforms the CE-trained LSTM that predicts 2,000 CD states.
Furthermore, although both systems are comparable in terms of the number of
parameters, the CTC-trained model is about $4\times$ faster than the CE-trained
baseline. Sequence discriminative training with the sMBR
criterion~\cite{Kingsbury:2009, SakVinyalsHeigold14} further improves system
performance by 20\% relative to the CTC-trained sytem.

In order to reduce memory consumption further, we compress our acoustic models
using projection layers that sit between the outputs of an LSTM layer and both
the recurrent and non-recurrent inputs to same and subsequent
layers~\cite{SakSeniorBeaufays14}.  Of crucial importance, however, is that when
a significant rank reduction is applied, it is not sufficient to simply
initialize the projection layer's weight matrix randomly for training with the
CTC criterion.  Instead we use the larger `uncompressed' model without the
projection layer and jointly factorize its recurrent and (non-recurrent)
inter-layer weight matrices at each hidden layer using a form of singular value
decomposition to determine a shared projection layer.  This process yields an
initialization that results in stable convergence as described in detail
in~\cite{prabhavalkar2016}. In our system, we introduce projection matrices of
rank 100 for the first four layers, and a projection matrix of rank 200 for the
fifth hidden layer.  Following SVD compression, we once again train the system
to optimize the CTC criterion, followed by discriminative sequence training with
the sMBR criterion. As can be seen in Table~\ref{tbl:am-results}, the proposed
compression technique allows us to compress the AM by about $3\times$.

Finally, we note that adapting the AM using a set of 1M anonymized
hand-transcribed utterances from the domain of open-ended dictation (processed to
generate multi-style training as described in Section~\ref{sec:mtr}) results in
a further 12.8\% relative improvement over the SVD compressed models. The
combination of all of these techniques allows us to significantly improve
performance over the baseline system.  For completeness, we also trained a
DNN system with topology described in ~\cite{lei2013}.  As expected, this 2,000
CD state DNN performed significantly worse than all of the LSTMs in
Table~\ref{tbl:am-results}.

For reference, we also present results obtained using a
much larger `server-sized' CTC model, which predicts 9287 CD phones (plus
``blank''), but is evaluated with the same LM and decoder graph as our other
systems, which serves as a sort of upperbound performance on this
task\footnote{This model uses 80-dimensional filterbank features in the
frontend, since this resulted in slightly improved performance. Frame
stacking and frame skipping are as in the CI LSTM CTC model.}.

\subsection{Efficient Representation and Fast Execution}
\label{sec:tinyfast}
Since the 11.9 MB floating point neural network acoustic model described above
consumes a significant chunk of the memory and processing-time, we quantize the
model parameters (i.e. weights, biases) into a more compact 8-bit integer-based
representation.  This quantization has an immediate impact on the memory usage,
reducing the acoustic model's footprint to a fourth of the original size.  The
final footprint of our AM is 3 MB as shown in Table~\ref{tbl:am-results}.  Using
8-bit integers also has the advantage that we can also achieve 8-way parallelism
in many matrix operations on most mobile platforms.

Although we could have applied a number of compression
schemes~\cite{bovik2005,gersho1992}, with simplicity and performance in mind,
and validated by previous work~\cite{vanhoucke2011}, we adopt a uniform linear
quantizer that assumes a uniform distribution of the values within a given
range. First, we find the minimum and maximum values of the original
parameters. We then use a simple mapping formula which determines a scaling
factor that when multiplied by the parameters spreads the values evenly in the
smaller precision scale, thus obtaining a quantized version of the original
parameters. The inverse operation is used when converting a quantized value back
to its 32-bit floating point equivalent.

During neural network inference, we operate in 8-bit integers everywhere except
in the activation functions and the final output of the network, which remain in
floating point precision (by converting between quantized 8-bit values and their
32-bit equivalents as needed). Our quantization scheme and the inference
computation approach provides a $2\times$ speed-up in evaluating our acoustic
models as compared to the unquantized model, with only a small performance
degredation (compare `adaptation' vs. `quantization' in
Table~\ref{tbl:am-results}).


%% file: lm.tex
\section{On-device Language Modelling}
\label{sec:lm}

In this work, we focus on building a compact language model for the domains of
dictation and voice commands. To maintain a small system footprint, we train a
single model for both domains. As described in Section~\ref{sec:baseline}, we
limit the vocabulary size to 64K. Our language models are trained using
unsupervised speech logs from the dictation domain ($\sim$100M utterances) and
voice commands domain ($\sim$2M utterances). The voice command utterances were
extracted by filtering general voice search queries through the grammars usually
used to parse voice commands at runtime. Those queries that parsed were added to
the training set.  A Katz-smoothed 5-gram LM is then trained and entropy-based
pruning is employed to shrink the LM to the sizes described in
Section~\ref{sec:baseline}.

In addition to the dictation test set described in Section~\ref{sec:am}, in this
section we present results on a voice commands test set. This set includes
utterances from 3 types of commands: Device ($\sim$2K utterances) - which includes
commands for device control (e.g., ``Turn up volume"), Planning
($\sim$9K utterances) - consisting of utterances relevant to planning calendar
events (e.g., ``Set an alarm at 6 p.m."), and Communication ($\sim$8K
utterances) with utterances relevant to chat messaging,
emails, or making phone calls. The Communication set, also includes some
open-ended dictation corresponding to the message (e.g. ``Text Jacob,
I'm running 10 minutes late, can we reschedule?").

All results in this section are evaluated using the quantized LSTM CI CTC
acoustic model described in Section~\ref{sec:am}, thus allowing us to focus on
the impact of the LM.

In order to build a single LM to use across both dictation and command domains,
we explore different interpolation techniques.
As our baseline, we consider a
linearly interpolated LM with interpolation weights estimated over a separate
held-out development set sampled from speech logs. We compare performance
obtained from the baseline system to a Bayesian interpolated
LM~\cite{AllauzenRiley11}, where voice commands and dictation are each
represented as a unique task and the corresponding task priors are determined by
sweeping parameters on a held-out development set to minimize word error rates
rather than setting these based on the log counts.


Our results are presented in Table~\ref{lmresults}.  The first two rows of the
table highlight the utility of Bayesian interpolation over linear interpolation
for both domains. The decoder graph used to produce these
results was constructed with a single large language model, and therefore
rescoring on-the-fly was not used. The third row of Table~\ref{lmresults} shows
the effects of on-the-fly rescoring on WER. Whereas the fully composed decoder
graph is an unacceptable 29 MB, breaking
them down into first-pass and on-the-fly rescoring models yields a 8.3 MB
decoder graph and a 6.8 MB rescoring LM (with LOUDS
compression~\cite{sorensen2011}).

\begin{table}
\begin{center}
\begin{tabular}{| c | c | c | c |}
\hline
LM Setup&Dictation WER&Commands WER\\
\hline
Linear Interpolation & 12.9 & 10.0\\
\hline
Bayesian Interpolation & 12.3 & 8.9\\
\hline
Bayesian + Rescoring & 13.5 & 9.0\\
\hline
\end{tabular}
\end{center}
\caption{Word Error Rates (\%) on an open-ended dictation domain and the
commands domain.}
\label{lmresults}
\end{table}

%% file: decoder.tex
\section{Decoder}
\label{sec:decoder}

\begin{figure}
  \centering
  \includegraphics[width=0.44\textwidth]{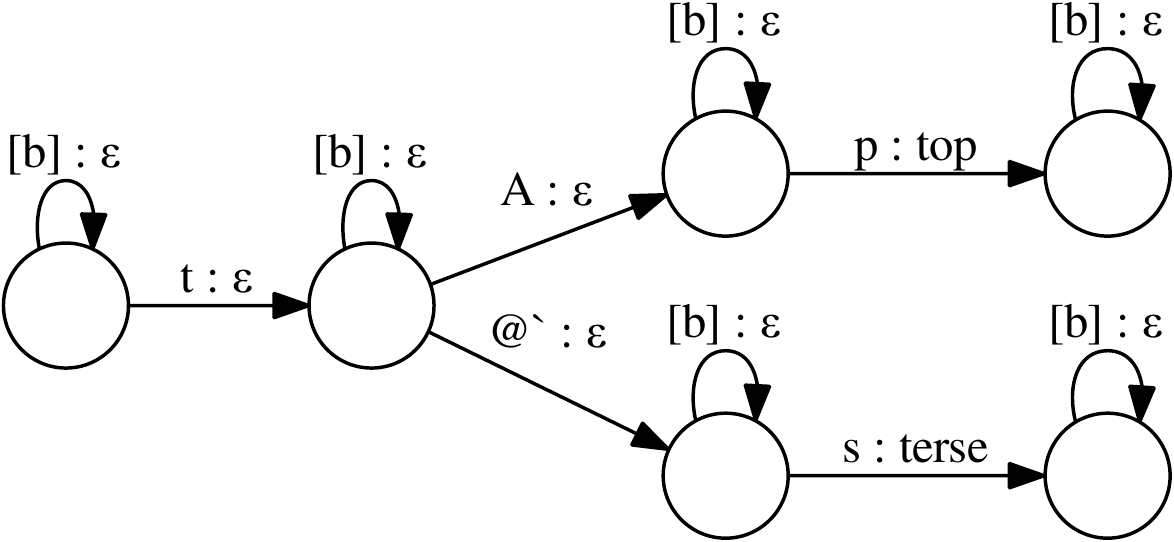}
  \caption{Example of a part of a decoder graph with \textit{blank} labels $[b]$.}
  \label{fig:decodergraph}
\end{figure}

In this section, we describe our decoder setup and a modification thereof that
takes advantage of CTC's simple topology. In contrast to a conventional 3-state
HMM structure, each phoneme is
represented by a single AM output state in combination with a generic
\textit{blank} (or ``non-perceiving") state. An FST-based decoder graph for
the CTC model is created by the usual construction and composition of
lexicon and LM transducers \cite{mohri2008}. We do not require a
context-dependency transducer, since we use context-independent phone models.
Self-loop transitions are added to each state for the \textit{blank} label.
An example is shown in Figure~\ref{fig:decodergraph}.

We use an FST-based decoder with optimizations for CTC models in terms of both
computation time and memory usage. By applying the \textit{blank} self-loop
transitions in the decoder, we can avoid adding them explicitly as arcs in the
decoder graph. Furthermore, the dynamic expansion of HMM state sequences used
in our generic FST-based decoder can be removed, which allows for a more
compact search space in memory and a simpler search hypothesis expansion
procedure.

%% file: context.tex
\section{Personalization}
\label{sec:context}

Our final set of experiments highlight the advantages of integrating personal
information into the language model. These experiments are aimed at determining
the impact of incorporating device-specific information (e.g., the user's list
of contact names) on the word error rate for individual users. We experiment
with two test sets related to contact name recognition.  The first is the 8K
utterance Communication test set described in Section~\ref{sec:lm}, containing
contact names in the context of messages, e.g., ``Text Jacob, $\ldots$''.  The
second set consists of 1.2K utterances containing \emph{only} contact names.
This second set is representative of the utterances that might follow a
text-to-speech (TTS) prompt such as: ``Which Jacob?'' or perhaps a more general
prompt such as ``Who would you like to email?''.  The number of candidate
contacts injected will depend on whether the TTS prompt is requesting
disambiguation or just any name from the contact list.  In either context, we
can perform the additional step of using on-the- fly rescoring as
in~\cite{hall2015} to bias the language model towards recognizing only these
contact names.

Given the lexical limits of the language model described above, it is unlikely
that the recognizer will be able to handle the long tail of contact names as is.
This motivates the incorporation of dynamic classes into our language model.  In
the general spirit of class-based LMs, and following the work of Aleksic
et. al.~\cite{aleksicIcassp2015} we annotate our training data with a
special~\emph{\$CONTACTS} symbol in place of contact names and train a language
model that includes this placeholder token. At run-time we inject a small FST
representing the user's personal contacts into the decoder graph at these
locations.  It should be noted that this is particularly simple in our system as
our AM uses context-independent phonemes.

In order to generate pronunciations for contacts we train a LSTM-based
grapheme-to-phoneme (G2P) model on human transcribed word-pronunciation pairs.
The G2P problem is treated as a sequence transcription task as described
in~\cite{lstm_g2p}. The LSTM-G2P system consists of four LSTM layers with 64
cells in each layer, and is trained to optimize the CTC objective function.  The
LSTM-G2P performs better in terms of word accuracy compared to traditional
joint-sequence models represented as finite state transducers (FSTs) (a detailed
comparison can be found in ~\cite{lstm_g2p}). More importantly, the LSTM-G2P is
considerably smaller in size compared to the FST implementation, 500 KB vs. 70
MB, making it an ideal solution for on-device pronunciation generation.

\begin{table}
\begin{center}
\begin{tabular}{|c|c|c|c|}
\hline
& Communication WER& Names WER & RT50\\
\hline
No Contacts & 13.7 & 70.3 & 0.14\\
\hline
2 Contacts & 9.0 & 30.0 & - \\
+ biasing &  -  & 12.8 & - \\
\hline
50 Contacts & 9.2 & 38.2 & - \\
+ biasing  & -    & 17.7 &  0.17\\
\hline
\end{tabular}
\end{center}
\caption{Impact of contact injection and biasing on WER and latency.}
\label{contacts}
\end{table}

Table~\ref{contacts} summarizes our results on the two contact test sets.  For
each utterance recognized, $N$ contacts are injected into the decoder graph.  If
the transcript does indeed contain a contact name, one of these $N$ is the
correct contact.  For the set containing only contact names, we additionaly
evaluate performance obtained using on-the-fly biasing~\cite{hall2015} towards
contact names.

Unsurprisingly, adding in personal contact names has a significant impact on
WER, since many of the terms in these test sets are out-of-vocabulary items.  In
contexts when a single contact name is the expected user-response, these results
indicate that biasing recognition towards the unigram \emph{\$CONTACTS} can
yield dramatic improvements, especially if the set of candidate names can be
whittled down to just two, as is often the case when disambiguating between
contacts (``Do you mean John Smith or John Snow?'').  While in practice one can
often precompute these graphs, we also show here that median RT factors are not
significantly affected even when 50 pronunciations are compiled and injected
on-the-fly in the system.

%% file: footprint.tex
\section{System Footprint}
\label{sec:footprint}
We present the sizes of the various components in our overall system
architecture in Table~\ref{tbl:sizes}. Using a combination of SVD-based
compression and quantization, along with a compact first-pass decoding strategy
and on-the-fly rescoring with a larger LM, we can build a system that is about
20.3 MB in size, without compromising accuracy or latency.
\begin{table}
  \centering
  \begin{tabular}{| c | c |}
    \hline
    Component & Size \\
    \hline
    \hline
    Acoustic Model & 3.0 MB \\
    \hline
    Decoder Graph & 8.3 MB \\
    \hline
    Rescoring LM & 6.8 MB \\
    \hline
    G2P Model & 497 KB \\
    \hline
    Text Normalizers & 1.1 MB \\
    \hline
    Endpointer & 22 KB \\
    \hline
    Personalization Components & 2.5 KB \\
    \hline
    \hline
    Total & 20.3 MB\\
    \hline
  \end{tabular}
  \caption{Size of various components in the overall system architecture.}
  \label{tbl:sizes}
\end{table}

%% file: conclusion.tex
\section{Conclusion}
\label{sec:conclusion}
We presented our design of a compact large vocabulary speech recognition system
that can run efficiently on mobile devices, accurately and with low latency.
This is achieved by using a CTC-based LSTM acoustic model which predicts
context-independent phones and is compressed to \emph{a tenth of its original
size} using a combination of SVD-based
compression~\cite{XueLiGong13, prabhavalkar2016} and quantization.

In order to support the domains of both open-ended dictation and voice
commands in a single language model we use a form of Bayesian interpolation.
Language model personalization is achieved through a combination of vocabulary
injection and on-the-fly language model biasing~\cite{aleksicIcassp2015,
hall2015}.

For efficient decoding, we use a on-the-fly rescoring strategy following
~\cite{lei2013} with additional optimizations for CTC models which reduce
computation and memory usage. The combination of these techniques allows us to
build a system which runs $7\times$ faster than real-time on a Nexus 5, with a
total system footprint of 20.3 MB.

%% file: acknowledgements.tex
\section{Acknowledgements}
\label{sec:acknowledgements}
The authors would like to thank our colleagues: Johan Schalkwyk, Chris Thornton,
Petar Aleksic, and Peter Brinkmann, for helpful research discussions and support
for the implementation.